\pdfoutput=1

\documentclass[11pt]{article}

\usepackage[final]{acl}

\usepackage{times}
\usepackage{latexsym}
\usepackage{graphicx}
\usepackage{latexsym}
\usepackage{supertabular}
\usepackage{stfloats}
\usepackage{tabularx}
\usepackage{multirow}
\usepackage{makecell}
\usepackage{booktabs}
\usepackage{amsmath}
\usepackage{amsfonts}
\usepackage{amssymb}
\usepackage{enumerate}
\usepackage{algorithm}
\usepackage{algorithmic}
\usepackage{bbm}
\usepackage{pifont}
\usepackage{colortbl}
\usepackage{xspace}
\usepackage{inconsolata}
\usepackage{subfigure}
\usepackage{caption}
\usepackage{subcaption}
\usepackage{enumitem}

\usepackage{float}  
\usepackage{subfigure}  
\usepackage{listings}
\usepackage{color}
\usepackage{wrapfig}
\usepackage{url}

\definecolor{dkgreen}{rgb}{0,0.6,0}
\definecolor{gray}{rgb}{0.5,0.5,0.5}
\definecolor{mauve}{rgb}{0.58,0,0.82}

\usepackage[T1]{fontenc}

\usepackage[utf8]{inputenc}

\usepackage{microtype}

\usepackage{inconsolata}

\usepackage{graphicx}

\title{CM-Align: Consistency-based Multilingual Alignment \\for Large Language Models}
\author{Xue Zhang\textsuperscript{1,2}\thanks{ \ This work was done during the internship at Pattern Recognition Center, WeChat AI, Tencent Inc, China.}, Yunlong Liang\textsuperscript{3}, 
Fandong Meng\textsuperscript{3}, Songming           Zhang\textsuperscript{1,2},\\
\textbf{Yufeng Chen\textsuperscript{1,2}\thanks{ \ \ Yufeng Chen is the corresponding author.}},
\textbf{Jinan Xu\textsuperscript{1,2}}, \textbf{Jie Zhou\textsuperscript{3}} \\
\textsuperscript{1}Key Laboratory of Big Data \& Artificial Intelligence in Transportation,\\Beijing Jiaotong University, Ministry of Education \\
\textsuperscript{2}School of Computer Science and Technology, Beijing Jiaotong University, Beijing, China \\
\textsuperscript{3}Pattern Recognition Center, WeChat AI, Tencent Inc, China \\
\texttt{\{\text{zhang\_xue},smzhang22,chenyf,jaxu\}@bjtu.edu.cn}}

\begin{document}
\maketitle
\begin{abstract}
Current large language models (LLMs) generally show a significant performance gap in alignment between English and other languages.
To bridge this gap, existing research typically leverages the model's responses in English as a reference to select the best/worst responses in other languages, which are then used for Direct Preference Optimization (DPO) training.
However, we argue that there are two limitations in the current methods that result in noisy multilingual preference data and further limited alignment performance:
1) Not all English responses are of high quality, and using a response with low quality may mislead the alignment for other languages.
2) Current methods usually use biased or heuristic approaches to construct multilingual preference pairs.
To address these limitations, we design a consistency-based data selection method to construct high-quality multilingual preference data for improving multilingual alignment (CM-Align).
Specifically, our method includes two parts: consistency-guided English reference selection and cross-lingual consistency-based multilingual preference data construction.
Experimental results on three LLMs and three common tasks demonstrate the effectiveness and superiority of our method, which further indicates the necessity of constructing high-quality preference data.

\end{abstract}

\section{Introduction}
Although existing large language models (LLMs) generally support multiple languages, the performance across different languages is heavily imbalanced \cite{qin2024multilinguallargelanguagemodel, xu2025survey, zhang-etal-2025-less} and typically shows significantly better general ability in English compared to other languages.
The main reason is that English, as the most popular language, benefits from abundant high-quality instruction tuning and preference data \cite{huang2025surveylargelanguagemodels}.
However, collecting such high-quality multilingual datasets is challenging due to the expensive human annotation and inevitable translation errors \cite{chen-etal-2024-breaking}.
To address this problem, existing research \cite{she-etal-2024-mapo, yang2025languageimbalancedrivenrewarding, yang2025implicitcrosslingualrewardingefficient} uses the English responses from the model itself as the reference to select the \textit{chosen}/\textit{rejected} pairs for multilingual data and then utilizes Direct Preference Optimization (\citealt{rafailov2024directpreferenceoptimizationlanguage}, DPO) to achieve multilingual alignment.

\begin{figure}[t]
    \centering
    \includegraphics[width=\linewidth]{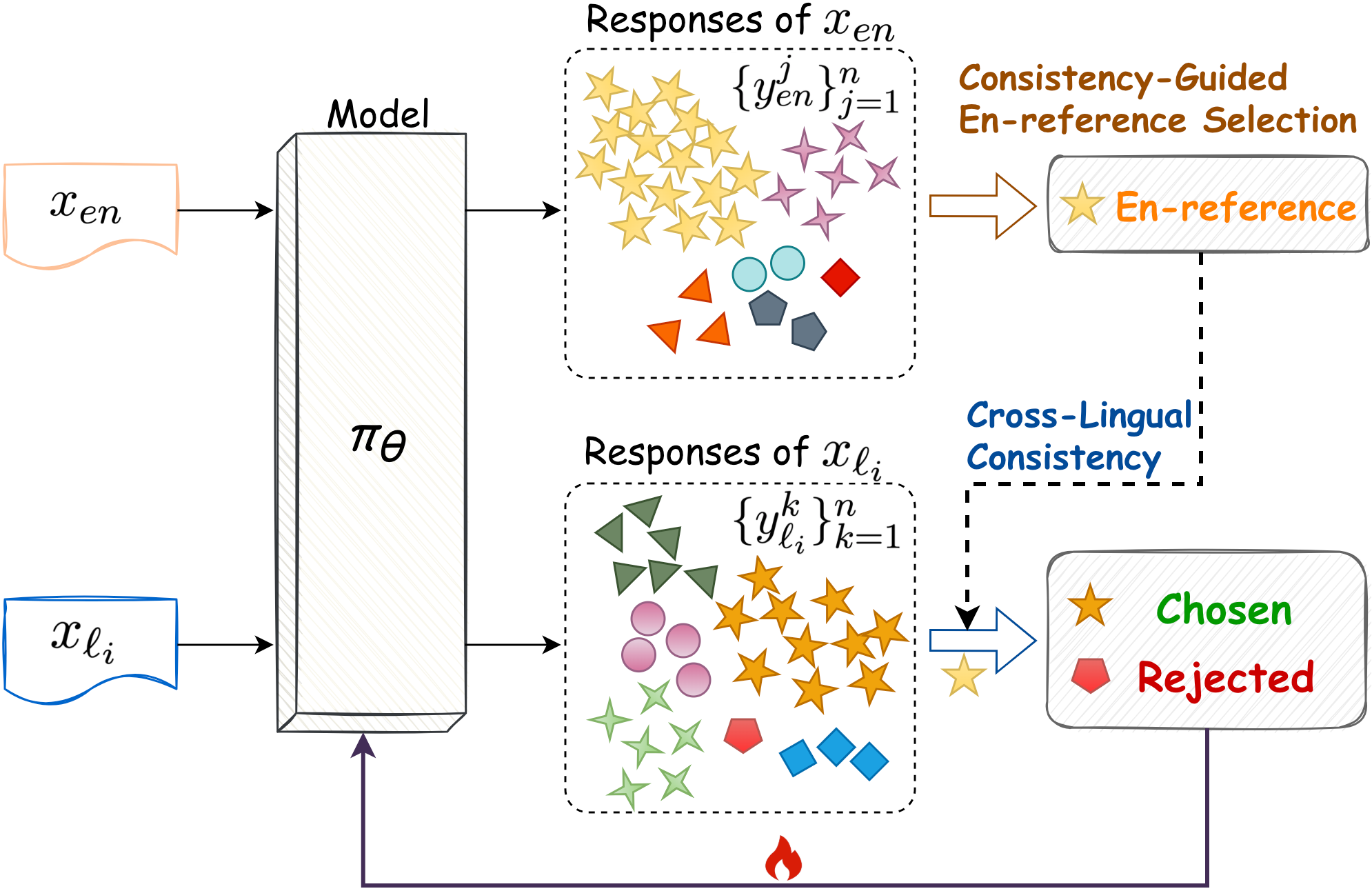}
    \caption{An overview of our method CM-Align. Given a question in different languages, we first request the model to generate multiple responses for each question. Then we select the most consistent one among all English responses as \textit{En-reference}, and calculate the cross-lingual consistency of multiple responses in other languages with \textit{En-reference} to determine the \textit{chosen}/\textit{rejected} example for DPO optimization.}
    \label{fig:method}
\end{figure}

However, previous work usually treats all English responses as equally valid and utilizes unreliable metrics or heuristic methods to construct multilingual preference pairs, which may introduce non-negligible noise to the preference data and result in limited alignment performance.
For example, MAPO \cite{she-etal-2024-mapo} uses the translation probability to find the most similar/dissimilar multilingual responses with the English responses to construct the \textit{chosen}/\textit{rejected} examples.
However, due to the strong semantic constraint, translation probability may not be a suitable metric for quality in scenarios like open-ended generation, math, and code snippets.
Additionally, LIDR \cite{yang2025languageimbalancedrivenrewarding} directly takes the translated answers from English to other languages as the \textit{chosen} examples and the original answers in other languages as the \textit{rejected} examples.
This heuristic design may introduce translation errors and translationese into the multilingual preference data.
Moreover, CLR \cite{yang2025implicitcrosslingualrewardingefficient} applies the self-rewarding method \cite{chen2024bootstrapping} to the multilingual scenario, which requires another aligned LLM to calculate implicit cross-lingual DPO rewards.
Although intuitive, English-centric LLMs may not be well-aligned in cross-lingual scenarios and thus offer inaccurate DPO rewards.

To solve these limitations, we design a consistency-based multilingual alignment method (CM-Align), which constructs high-quality multilingual preference data for DPO training (as shown in Figure \ref{fig:method}).
Our method improves the quality of multilingual preference data from two perspectives: \textbf{1)} consistency-guided English reference selection to find the reliable English anchor, and \textbf{2)} cross-lingual consistency-based preference data construction with task-specific metrics.
Specifically, for each prompt, we first sample multiple responses in English and select the one that is most consistent with others as \textit{En-reference}.
Then we choose the responses in other languages that are most consistent with \textit{En-reference} as the \textit{chosen} example and the most inconsistent response as the \textit{rejected} example.
Particularly, for different tasks, we design task-specific criteria to better measure the consistency between different responses.
Experimental results on Math, Code, and General Instruction Following tasks demonstrate the effectiveness and superiority of our method. 
Moreover, our method not only shows better alignment performance for in-domain languages but also exhibits excellent generalization for out-of-domain languages, which proves that different languages can learn collaboratively for LLM alignment.

In summary, the major contributions of this paper are as follows\footnote{The code is publicly available at \url{https://github.com/XZhang00/CM-Align}.}:
\begin{itemize}
    \item We propose a novel multilingual alignment method named CM-Align, which constructs high-quality multilingual preference data based on self-consistency and cross-lingual consistency.
    \item We design task-specific criteria for evaluating the consistency of any two responses for each prompt, which can be applied in other similar scenarios.
    \item We evaluate our method on three LLMs and three common tasks, proving the superiority and extensive application of our method.
    
\end{itemize}


\section{Related Work}

\paragraph{Reinforcement Learning from Human Feedback.} 
RLHF \cite{christiano2023deepreinforcementlearninghuman, ziegler2020finetuninglanguagemodelshuman, ouyang2022traininglanguagemodelsfollow} is the critical technique to align LLMs with human preferences and values.
Among multiple implementations, Direct Preference Optimization (\citealt{rafailov2024directpreferenceoptimizationlanguage}, DPO) streamlines the alignment process by directly optimizing the LLM policy from preference data, which needs fewer computational requirements and shows comparable effectiveness compared to classical RLHF methods.
Recently, some works \cite{azar2023generaltheoreticalparadigmunderstand, meng2024simposimplepreferenceoptimization, tang2024generalizedpreferenceoptimizationunified, hong2024orpomonolithicpreferenceoptimization, zhang2025aligndistiltokenlevellanguagemodel} modify the training objectives to improve the performance of DPO.
Additionally, other works \cite{deng2025moreimprovingllmalignment, li2025crowdselectsyntheticinstructiondata, xiao2025findingsweetspotpreference} focus on the data selection method to improve the training performance.
However, these methods only concentrate on the alignment for English without considering the challenges of multilingual scenarios.

\paragraph{Multilingual Alignment.} 
Currently, many works aim to align the multilingual ability of LLMs to English.
\citet{lai2023okapiinstructiontunedlargelanguage} translates English preference data into other languages, which is limited to translation errors.
\citet{wu-etal-2024-reuse} directly applies the reward model trained in the source language to other target languages.
\citet{dang-etal-2024-rlhf} focuses on exploring data coverage of different languages.
Additionally, some works \cite{she-etal-2024-mapo, yang2025languageimbalancedrivenrewarding, yang2025implicitcrosslingualrewardingefficient} try to utilize the dominant English to assist the construction of multilingual preference data.
MAPO \cite{she-etal-2024-mapo} adopts the NLLB model to calculate the translation probability scores of the answers in non-dominant languages and English, and chooses the answer with the highest/lowest score as the \textit{chosen}/\textit{rejected} example. 
LIDR \cite{yang2025languageimbalancedrivenrewarding} leverages the characteristics of translation, \textit{i.e.}, the answers translated with the model itself from English to non-dominant languages as \textit{chosen} examples and the original answers in non-dominant languages as \textit{rejected} examples.
CLR \cite{yang2025implicitcrosslingualrewardingefficient} adapts the self-rewarding method to the multilingual scenario, which needs another SFT model to calculate implicit cross-lingual rewards.
Although these methods have an initial attempt, we argue that the native translation rewards or cross-lingual rewards are unfaithful, further leading to low-quality preference data and underperformed multilingual alignment performance.
In this paper, we aim to select high-quality preference data to improve multilingual alignment.

\section{Methodology}
\subsection{Background}

RLHF provides a systematic framework for aligning language models with human preferences through a two-phase pipeline: rewarding modeling and policy optimization.

\paragraph{Preference-Driven Reward Modeling.}
Given a dataset with $N$ samples $\mathcal{D} = \{(x, y_w, y_l)_i\}_N$, where $x$ denotes the input prompt and $y_w/y_l$ represents the human-annotated preferred/dispreferred response, the reward model $r_\phi$ is trained using the Bradley-Terry preference model \citep{bradley1952rank}. The training objective minimizes the contrastive loss:
\begin{align} \label{eq:reward_modeling}
    &\mathcal{L}_{\rm RM}(\phi)= \nonumber \\
    &-\!\!\!\!\mathop{\mathbb{E}}_{(x,y_w,y_l) \sim \mathcal{D}} \Big[\log \sigma (r_{\phi}(x,y_w) - r_{\phi}(x,y_l)) \Big],
\end{align}
where $\sigma(\cdot)$ is the sigmoid function that converts reward differences into preference probabilities.

\paragraph{Policy Optimization.}
The learned reward function $r_\phi$ then guides policy optimization through RL algorithms like PPO \citep{schulman2017proximal}. The objective balances reward maximization against policy divergence:
\begin{align} \label{eq:rlhf_obj}
    &\mathcal{J}_{\rm RLHF}(\theta)= \nonumber \\
    &\max_{\theta}\mathop{\mathbb{E}}_{\substack{x \sim \mathcal{D} \\ y \sim \pi_{\theta}(\cdot|x)}} \Big[ r_{\phi}(x,y) - \beta \log \frac{\pi_{\theta}(y|x)}{\pi_{\rm ref}(y|x)} \Big],
\end{align}
here $\beta > 0$ controls the Kullback-Leibler (KL) divergence \citep{kullback1951information} that prevents excessive deviation from the reference policy $\pi_{\mathrm{ref}}$. While effective, this approach requires complex RL implementations and exhibits sensitivity to reward miscalibration.

\paragraph{Direct Preference Optimization.}
As an alternative, DPO \cite{rafailov2024directpreferenceoptimizationlanguage} eliminates the separate reward modeling phase through implicit reward parameterization. By establishing a mapping between reward functions and optimal policies, it derives a closed-form solution:

\begin{equation}\label{eq:init_dpo_reward}
r_\theta(x,y) = \beta \log\frac{\pi_\theta(y|x)}{\pi_{\mathrm{ref}}(y|x)} + \beta \log Z(x),
\end{equation}
where $Z(x)$ denotes the partition function and independent to $y$ that normalizes the reward distribution.
Then the training objective of DPO is derived by substituting Eq. (\ref{eq:init_dpo_reward}) into Eq. (\ref{eq:reward_modeling}):
\begin{align} \label{eq:dpo_loss}
    &\mathcal{L}_{\rm DPO}(\theta)=-{\mathbb{E}}_{(x,y_w,y_l)\sim \mathcal{D}}\Big[ \nonumber \\
    &\log \sigma \Big(\beta\log\frac{\pi_{\theta}(y_w|x)}{\pi_{\rm ref}(y_w|x)} - \beta\log\frac{\pi_{\theta}(y_l|x)}{\pi_{\rm ref}(y_l|x)}\Big) \Big].
\end{align}

To sum up, DPO streamlines the alignment process by directly optimizing for human preferences, offering superior computational efficiency and stability compared to traditional RLHF approaches, without compromising on alignment quality.


\subsection{CM-Align}
Our approach includes two components: consistency-guided English reference selection and cross-lingual consistency-based construction of multilingual preference data.
The key idea is to select a high-quality English response as the reference to guide the choice of the preferred and dispreferred responses for other languages.

Given the model $\pi_\theta$ and a multilingual prompt set $\mathcal{Q} = \{x_{en}, x_{\ell_1}, ..., x_{\ell_k}\}$, where $x_{en}$ denotes an English prompt and $x_{\ell_i}$ represent the parallel prompt in language $\ell_i$, we first request $\pi_\theta$ to generate $n$ responses for each prompt in $\mathcal{Q}$ as the candidate samples.

\paragraph{Consistency-Guided English Reference Selection.}
Given one English prompt $x_{en}$ and the generated $n$ candidate responses $\{y_{en}^j\}_{j=1}^n$ through stochastic sampling from the target model $\pi_\theta$, we select the most consistent response $y_{en}^*$ with each other candidate response as \textit{En-reference}, which is generally a more reliable anchor \cite{wang2023selfconsistencyimproveschainthought}.

To judge the consistency of any two responses, we design task-specific criteria for three common types of tasks (Math, Code, and General Instruction Following).
Specifically, for responses to Math prompts, we first extract the final numerical answer with regular expressions\footnote{The regular expressions are listed in Appendix \ref{sec:appendix-regx_math}.} and conduct majority voting\footnote{When there are several winners for the math task, we randomly select one as \textit{En-reference}.} to obtain the \textit{En-reference}.
For the responses to Code prompts, we first extract the code snippet and normalize\footnote{The codes for normalization are listed in Appendix \ref{sec:appendix-codes_for_code}.} the code snippet, including comment removal, variable anonymization, and code format standardization.
Then, given two normalized code snippets, we calculate the weighted average of CodeBLEU \cite{ren2020codebleumethodautomaticevaluation} and CodeBERTScore \cite{zhou2023codebertscoreevaluatingcodegeneration} as the consistency score of any two code responses.
Formally, $\mathrm{Cons}_{\text{code}}({c_{en}^1}, {c_{en}^2})=\alpha \cdot \mathrm{CodeBLEU} + (1-\alpha) \cdot \mathrm{CodeBERTScore}$, where $c_{en}^1$ is the code snippet after the normalization procedures originally extracted from $y_{en}^1$ and $\alpha$ is the hyperparameter to control the weight of CodeBLEU and CodeBERTScore.
For the responses to General Instruction Following (GIF) prompts, we first encode each response with the \texttt{gte-large-en-v1.5}\footnote{The \texttt{gte-large-en-v1.5} model is an instruction-tuned multi-lingual embedding model \cite{li2023towards} released by Alibaba-NLP at \url{https://huggingface.co/Alibaba-NLP/gte-large-en-v1.5}.} model to an embedding and then compute the cosine similarity of two embeddings as the consistency score.
Formally, $\mathrm{Cons}_{\text{GIF}}({y_{en}^1}, {y_{en}^2})=\mathrm{cos}(emb(y_{en}^1), emb(y_{en}^2))$, where $emb(y_{en}^1)$ is the encoded embedding of $y_{en}^1$.

According to these criteria, we can calculate the consistency score for any two responses, and select the most consistent response $y_{en}^*$ as \textit{En-reference} for each prompt $x_{en}$.


\paragraph{Cross-Lingual Consistency-based Preference Data Construction.}
For each multilingual prompt $x_{\ell_i}$, we first generate $n$ responses $\{y_{\ell_i}^k\}_{k=1}^n$ and then compute the cross-lingual consistency score relative to $y_{en}^*$ for each $y_{\ell_i}^k$, \textit{i.e.}, $\mathrm{CL\text{-}Cons}(y_{\ell_i}^k, y_{en}^*)$.
The cross-lingual consistency score is also task-specific.
Specifically, for responses to Math prompts, we also first extract the final numerical answer with regular expressions and judge whether the answer is consistent with the extracted answer of $y_{en}^*$.
For responses to Code prompts, we first extract the code snippet from $y_{\ell_i}^k$ and normalize it as $c_{\ell_i}^k$. 
Then we compute the cross-lingual consistency scores $\mathrm{CL\text{-}Cons}_{\text{code}}(y_{\ell_i}^k, y_{en}^*) = \mathrm{Cons}_{\text{code}}({c_{\ell_i}^k}, {c_{en}^*})$, where ${c_{en}^*}$ is the normalized code extracted from $y_{en}^*$.
For the responses to multilingual GIF prompts, we first encode each response with the \texttt{gte-multilingual-base}\footnote{The \texttt{gte-multilingual-base} model achieves state-of-the-art (SOTA) results in multilingual retrieval tasks and multi-task representation model evaluations when compared to models of similar size \cite{zhang2024mgte} released by Alibaba-NLP at \url{https://huggingface.co/Alibaba-NLP/gte-multilingual-base}.} model to an embedding and then compute the cosine similarity between the embedding in other languages and the embedding of $y_{en}^*$ as the cross-lingual consistency score, \textit{i.e.},  $\mathrm{CL\text{-}Cons}_{\text{GIF}}({y_{\ell_i}^k}, {y_{en}^*})=\mathrm{cos}(emb(y_{\ell_i}^k), emb(y_{en}^*))$.


Finally, we select the response with the highest cross-lingual consistency score as the \textit{chosen} example $y^{{w}}_{\ell_i}$, and the lowest as the \textit{rejected} example $y^{{l}}_{\ell_i}$, \textit{i.e.}, the multilingual DPO training pairs\footnote{The constructed data includes the English preference data.} are constructed as:

\begin{align}
y^{{w}}_{\ell_i} = \arg\max_{y_{\ell_i}^k} \mathrm{CL\text{-}Cons}(y_{\ell_i}^k, y_{en}^*), \\
y^{{l}}_{\ell_i} = \arg\min_{y_{\ell_i}^k} \mathrm{CL\text{-}Cons}(y_{\ell_i}^k, y_{en}^*)).
\end{align}

Particularly, we filter out samples that cannot be used to construct \textit{chosen} and \textit{rejected} examples, such as when all the extracted numerical answers or normalized codes are consistent.
The filtering strategy guarantees the difference and rationality of DPO training pairs.

\paragraph{Multilingual Preference Optimization.}
Given the multilingual preference dataset $\mathcal{D}_M = \{(x_{\ell_i}, y^{{w}}_{\ell_i}, y^{{l}}_{\ell_i})\}$, the DPO objective is formulated as:

\begin{align} \label{eq:dpo_loss_m}
    &\mathcal{L}_{\rm DPO}(\theta)=-{\mathbb{E}}_{(x_{\ell_i}, y^{{w}}_{\ell_i}, y^{{l}}_{\ell_i})\sim \mathcal{D}_M}\Big[ \nonumber \\
    &\log \sigma \Big(\beta\log\frac{\pi_{\theta}(y^{{w}}_{\ell_i}|x_{\ell_i})}{\pi_{\rm ref}(y^{{w}}_{\ell_i}|x_{\ell_i})} - \beta\log\frac{\pi_{\theta}(y^{{l}}_{\ell_i}|x_{\ell_i})}{\pi_{\rm ref}(y^{{l}}_{\ell_i}|x_{\ell_i})}\Big) \Big].
\end{align}
Additionally, a negative log-likelihood (NLL) loss for the \textit{chosen} examples is incorporated into the vanilla DPO \cite{rafailov2024directpreferenceoptimizationlanguage} to improve alignment performance. The NLL objective is as follows:

\begin{equation}
    \mathcal{L}_{\rm NLL} = -{\mathbb{E}}_{(x_{\ell_i}, y^{{w}}_{\ell_i})\sim \mathcal{D}_M} \Big[ \log \frac{\pi_{\theta}(y^{{w}}_{\ell_i}|x_{\ell_i})}{|y^{{w}}_{\ell_i}|} \Big].
\end{equation}

Overall, the training objective $\mathcal{L}$ is: $\mathcal{L} = \mathcal{L}_{\rm DPO} + \gamma \mathcal{L}_{\rm NLL}$, where $\gamma$ is a hyper-parameter that controls the weight of $\mathcal{L}_{\rm NLL}$.

\section{Experiments}

\subsection{Experimental Setup}

In our experiments, we select three multilingual LLMs of different scales: Llama-3.2-3B-Instruct \cite{grattafiori2024llama3herdmodels}, Qwen2.5-3B-Instruct \cite{qwen2025qwen25technicalreport}, and Llama-3-8B-Instruct \cite{llama3modelcard}.
The three models are all English-centric, \textit{i.e.}, exhibiting suboptimal performance on other languages.
Therefore, we aim to utilize the English proficiency to improve the ability of other languages.
We conduct main experiments on three common tasks: Math, Code, and General Instruction Following \cite{zhang2025dualspaceframeworkgeneralknowledge}.
The benchmark and training data for each task are as follows:

For Math, we use the MGSM \cite{shi2022languagemodelsmultilingualchainofthought} benchmark to evaluate the mathematical performance of different languages, which involves ten languages: English (\textit{en}), Chinese (\textit{zh}), Spanish (\textit{es}), French (\textit{fr}), Russian (\textit{ru}), Bengali (\textit{bn}), German (\textit{de}), Thai (\textit{th}), Swahili (\textit{sw}), and Japanese (\textit{ja}).
Then we randomly select 4.5K English questions (with the ground-truth labels) from MetaMath \cite{yu2024metamathbootstrapmathematicalquestions} and request \texttt{gpt-4o-2024-08-06} to translate\footnote{The translation prompts follow \url{https://github.com/OpenBMB/UltraLink/blob/main/multi-math/math_prompt.yaml}.} English questions to the four languages \textit{zh}, \textit{es}, \textit{fr}, and \textit{ru} as the training data.
And the other five languages \textit{bn}, \textit{de}, \textit{th}, \textit{sw}, and \textit{ja} are unseen languages for observing the out-of-domain generalization of each method.

For Code, we use the Multilingual HumanEval \cite{wang-etal-2024-ultralink} benchmark to evaluate the code generation ability of different languages, which includes five languages: \textit{en}, \textit{zh}, \textit{es}, \textit{fr},  and \textit{ru}. 
And we translate\footnote{The translation prompts follow \url{https://github.com/OpenBMB/UltraLink/blob/main/multi-code/code_prompt.yaml}.} the benchmark with \texttt{gpt-4o-2024-08-06} to \textit{bn} and \textit{de} as the out-of-domain languages.
For the training data, we randomly select 2K examples from the Python subset of Magicoder \cite{wei2024magicoderempoweringcodegeneration} and request \texttt{gpt-4o-2024-08-06} to translate English instructions to the four languages \textit{zh}, \textit{es}, \textit{fr}, and \textit{ru}.

For General Instruction Following, we use OMGEval \cite{liu2024omgevalopenmultilingualgenerative} as the evaluation benchmark, which is the multilingual version of AlpacaEval \cite{alpaca_eval} (including \textit{en}, \textit{zh}, \textit{es}, \textit{fr}, \textit{ru}, \textit{bn}, and \textit{de}) and has undergone the specific localization process and human check.
And we also set \textit{bn} and \textit{de} as the out-of-domain languages.
We select the \texttt{GPT3.5-turbo} model as the baseline comparison model.
For the training data, we randomly select 1.5K instructions from Alpagasus \cite{chen2024alpagasustrainingbetteralpaca} and request \texttt{gpt-4o-2024-08-06} to translate\footnote{The translation prompts follow \url{https://github.com/OpenBMB/UltraLink/blob/main/multi-sharegpt/sharegpt_prompt.yaml}.} English instructions to the four languages \textit{zh}, \textit{es}, \textit{fr}, and \textit{ru}.

In all experiments, we generate $n=30$ responses with temperature 0.5 and top-p 0.9 for each prompt.
And the weight of CodeBLEU is set to $\alpha=0.7$.
The other training details are listed in Table \ref{tab:train_config} of Appendix \ref{appendix:training_details}.


\begin{table*}[t]
    \centering
    \resizebox*{\linewidth}{!}{
    \begin{tabular}{l|ccccc|c|ccccc|c|c}
    \bottomrule
& \multicolumn{6}{c|}{\textbf{In-Domain Languages}} & \multicolumn{6}{c|}{\textbf{Out-of-Domain Languages}} & \\
\hline
\textbf{Methods}	&	\textbf{en}	&	\textbf{zh}	&	\textbf{es}	&	\textbf{fr}	&	\textbf{ru}	&	\textbf{\textit{ID-avg}}	&	\textbf{bn}	&	\textbf{de}	&	\textbf{th}	&	\textbf{sw}	&	\textbf{ja}	&	\textbf{\textit{OOD-avg}}	&	\textbf{\textit{All-avg}}	\\
\hline
\rowcolor{gray!30}
\textbf{Llama-3.2-3B-Instruct}	&	71.60 	&	59.60 	&	67.20 	&	60.80 	&	62.40 	&	64.32 	&	44.00 	&	63.20 	&	58.80 	&	51.20 	&	49.20 	&	53.28 	&	58.80 	\\
\hline
\multicolumn{14}{c}{\textbf{\textit{with label}}} \\
\hline
SFT-translation 	&	79.20 	&	62.80 	&	70.40 	&	66.80 	&	67.60 	&	69.36 	&	50.40 	&	65.20 	&	58.00 	&	52.80 	&	52.80 	&	55.84 	&	62.60 	\\
SFT-self-rejection 	&	76.00 	&	60.80 	&	72.00 	&	62.80 	&	67.60 	&	67.84 	&	50.00 	&	64.40 	&	\textbf{60.80} 	&	57.20 	&	51.20 	&	56.72 	&	62.28 	\\
CM-Align (Ours) 	&	\textbf{81.60} 	&	\textbf{65.60} 	&	\textbf{74.00} 	&	\textbf{69.60} 	&	\textbf{70.80} 	&	\textbf{72.32} 	&	\textbf{51.60} 	&	\textbf{68.80} 	&	59.20 	&	\textbf{58.40} 	&	\textbf{58.80} 	&	\textbf{59.36} 	&	\textbf{65.84} 	\\
\hline
\multicolumn{14}{c}{\textbf{\textit{label-free}}} \\
\hline
Random-selection	&	72.00 	&	60.80 	&	68.00 	&	60.00 	&	60.40 	&	64.24 	&	48.40 	&	61.20 	&	54.80 	&	52.40 	&	48.80 	&	53.12 	&	58.68 	\\
MAPO	&	74.40 	&	60.40 	&	66.80 	&	60.00 	&	64.80 	&	65.28 	&	51.20 	&	60.40 	&	58.00 	&	54.40 	&	51.60 	&	55.12 	&	60.20 	\\
LIDR	&	70.40 	&	55.60 	&	63.20 	&	56.40 	&	58.00 	&	60.72 	&	\textbf{51.60} 	&	61.60 	&	54.40 	&	50.00 	&	51.60 	&	53.84 	&	57.28 	\\	
CM-Align (Ours) 	&	\textbf{78.40} 	&	\textbf{63.60} 	&	\textbf{72.40} 	&	\textbf{68.40} 	&	\textbf{68.40} 	&	\textbf{70.24} 	&	50.80 	&	\textbf{67.20} 	&	\textbf{60.80} 	&	\textbf{61.20} 	&	\textbf{60.00} 	&	\textbf{60.00} 	&	\textbf{65.12} 	\\
\toprule
\bottomrule
\rowcolor{gray!25}
\textbf{Qwen2.5-3B-Instruct}	&	 81.60 &	72.00 &	69.60 &	65.20 &	68.00 &	71.28 &	31.60 &	66.80 &	59.60 &	3.20 &	52.80 &	42.80 	& 57.04 	\\
\hline
\multicolumn{14}{c}{\textbf{\textit{with label}}} \\
\hline
SFT-translation	&	\textbf{82.80} &	72.80 &	66.40 & 65.20 &	63.60 &	70.16 &	28.40 &	65.20 &	56.00 & 5.60 &	51.20  &	41.28 &	55.72 	\\
SFT-self-rejection	&	 78.80 &	70.00 &	68.00 &		64.40 &	74.00 &	71.04 &		35.60 &	66.80 &		58.40 &		5.20 &		56.00 &		44.40 &		57.72	\\
CM-Align (Ours)	&	\textbf{82.80} &	\textbf{73.60} &	\textbf{75.20} &	\textbf{69.60} &	\textbf{76.00} &	\textbf{75.44} &	\textbf{41.20} &	\textbf{69.60} &	\textbf{60.00} &	\textbf{8.00} &	\textbf{57.60} &	\textbf{47.28} &	\textbf{61.36} 	\\
\hline
\multicolumn{14}{c}{\textbf{\textit{label-free}}} \\
\hline

Random-selection &	79.60 &	73.60 &	66.80 &	64.80 &	68.00 &	70.56 &	36.80 &	65.60 &	57.60 &	6.00 &	56.00 &	44.40 &	57.48  	\\
MAPO	&	80.00 &	73.60 &	\textbf{72.40} &	65.60 &	72.00 &	72.72 &	37.20 &	\textbf{68.00} &	\textbf{64.80} &	6.40 &	\textbf{60.00} &	47.28 &	60.00 	\\
LIDR	&	 78.80 &	69.20 &	66.40 &	62.40 &	67.60 &	68.88 &	33.20 &	60.80 &	58.80 &	4.00 &	55.20 &	42.40 &	55.64 	\\
CM-Align (Ours)	&	 \textbf{80.80} &	\textbf{75.20} &	71.60 &	\textbf{68.00} &	\textbf{72.80} &	\textbf{73.68} &	\textbf{42.40} &	66.80 &	63.20 &	\textbf{8.00} &	58.40 &	\textbf{47.76} &	\textbf{60.72} 	\\
\toprule
\bottomrule
\rowcolor{gray!25}
\textbf{Llama-3-8B-Instruct}	&	75.60 	&	64.00 	&	63.60 	&	58.00 	&	63.60 	&	64.96 	&	46.00 	&	60.80 	&	58.40 	&	39.20 	&	50.00 	&	50.88 	&	57.92 	\\
\hline
\multicolumn{14}{c}{\textbf{\textit{with label}}} \\
\hline
SFT-translation	&	\textbf{80.40} 	&	59.60 	&	\textbf{73.20} 	&	\textbf{66.80} 	&	\textbf{68.80} 	&	\textbf{69.76} 	&	40.80 	&	65.20 	&	60.80 	&	39.20 	&	51.20 	&	51.44 	&	60.60 	\\
SFT-self-rejection	&	75.60 	&	62.40 	&	66.80 	&	58.40 	&	65.60 	&	65.76 	&	46.40 	&	62.00 	&	57.60 	&	34.40 	&	50.80 	&	50.24 	&	58.00 	\\
CM-Align (Ours)	&	77.20 	&	\textbf{64.80} 	&	72.00 	&	64.00 	&	67.60 	&	69.12 	&	\textbf{49.20} 	&	\textbf{66.80} 	&	\textbf{61.20} 	&	\textbf{40.80} 	&	\textbf{52.80} 	&	\textbf{54.16} 	&	\textbf{61.64} 	\\
\hline
\multicolumn{14}{c}{\textbf{\textit{label-free}}} \\
\hline

Random-selection	&	72.80 	&	55.20 	&	62.40 	&	58.40 	&	57.20 	&	61.20 	&	51.60 	&	63.60 	&	59.20 	&	38.00 	&	49.20 	&	52.32 	&	56.76 	\\
MAPO	&	\textbf{76.40} 	&	\textbf{66.80} 	&	65.60 	&	58.40 	&	59.60 	&	65.36 	&	\textbf{55.60} 	&	\textbf{65.60} 	&	56.80 	&	42.00 	&	\textbf{54.80} 	&	54.96 	&	60.16 	\\
LIDR	&	74.40 	&	64.00 	&	65.60 	&	56.00 	&	61.60 	&	64.32 	&	52.00 	&	47.60 	&	53.20 	&	32.00 	&	38.80 	&	44.72 	&	54.52 	\\
CM-Align (Ours)	&	74.80 	&	65.60 	&	\textbf{71.20} 	&	\textbf{62.40} 	&	\textbf{62.40} 	&	\textbf{67.28} 	&	49.60 	&	64.80 	&	\textbf{65.20} 	&	\textbf{42.40} 	&	53.60 	&	\textbf{55.12} 	&	\textbf{61.20} 	\\
    \toprule
    \end{tabular}
    }
    \caption{
        The accuracy (\%) results on the MGSM benchmark of the three models. ``\textit{\textbf{with label}}'' denotes conducting the \textit{En-selection} selection according to the ground-truth label, while ``\textit{\textbf{label-free}}'' means without access to the ground-truth label. ``\textit{\textbf{ID-avg}}/\textit{\textbf{OOD-avg}}'' is the average result of five In-Domain/Out-of-Domain languages and ``\textit{\textbf{All-avg}}'' is the average result of all ten languages. The result in \textbf{bold} means the best result in each setting.
    }
    \label{table:math-res}
\end{table*}

\subsection{Baselines}
\paragraph{SFT-translation.}
Only for Math, we request \texttt{gpt-4o-2024-08-06} to translate 4.5K English questions and answers to \textit{zh}, \textit{es}, \textit{fr}, and \textit{ru}.
And we conduct supervised fine-tuned (SFT) training to explore the alignment performance of translated data.
Particularly, the answer to each question in any language is accurate.

\paragraph{SFT-self-rejection.}
Only for Math, we request the model itself to generate 30 responses for each question and conduct rejection sampling for the SFT training, \textit{i.e.}, only select the accurate answer for each question according to the ground-truth label.

\paragraph{Random-selection.}
We randomly select the \textit{chosen}/\textit{rejected} responses for each prompt in any language.

\paragraph{MAPO.}
MAPO \cite{she-etal-2024-mapo} calculates the translation probability between the responses in English and other languages with the \texttt{nllb-200-distilled-600M}\footnote{\url{https://huggingface.co/facebook/nllb-200-distilled-600M}} model, where the responses with the highest/lowest scores are selected as the \textit{chosen}/\textit{rejected} examples.

\paragraph{LIDR.}
LIDR \cite{yang2025languageimbalancedrivenrewarding} leverages the characteristics of translation to construct preference data, \textit{i.e.}, the answers translated by the model itself from English to non-dominant languages as \textit{chosen} examples and the original answers in non-dominant languages as \textit{rejected} examples.
For English, the original generated answers are \textit{chosen} examples, and translated from answers in non-dominant languages to English are \textit{rejected} examples.

\begin{table*}[t]
    \centering
    \resizebox*{0.9\linewidth}{!}{
    \begin{tabular}{l|ccccc|c|cc|c|c}
    \bottomrule
 & \multicolumn{6}{c|}{\textbf{In-Domain Languages}} & \multicolumn{3}{c|}{\textbf{Out-of-Domain Languages}} & \\
\hline
\textbf{\textit{label-free}} \textbf{Methods}	&	\textbf{en}	&	\textbf{zh}	&	\textbf{es}	&	\textbf{fr}	&	\textbf{ru}	&	\quad \textbf{\textit{ID-avg}} \quad	&	\textbf{bn}	&	\textbf{de}	& 	\textbf{\textit{OOD-avg}}	& \quad	\textbf{\textit{All-avg}} \quad	\\
\hline
\rowcolor{gray!30}
\textbf{Llama-3.2-3B-Instruct}	&	35.37 & 30.89 & 	21.95 & 	23.37 & 	32.32 & 	28.78 & 	35.77 & 26.42 & 31.10 & 	29.44 	\\
\hline
Random-selection & 33.54	&	31.91	&	24.39	&	22.15	&	32.93	&	28.98	&	35.77	&	23.17	&	29.47	&		29.12	\\
MAPO	&	46.34	&		36.99	&	33.33	&	34.55	&	38.21	&	37.89	&	38.41	&	36.38	&	37.40	&	37.75 \\
LIDR	&	22.36	&	22.36	&	5.89	&	13.01	&	22.15	&	17.15 	&	17.28	&		9.76	&	13.52	&		16.11	\\	
CM-Align (Ours) &	\textbf{53.66}	&		\textbf{52.03}	&	\textbf{41.67}	&	\textbf{37.20}	&	\textbf{44.92}	&	\textbf{45.89}	&		\textbf{46.14}	&	\textbf{56.10}	&	\textbf{51.12}	&	\textbf{47.39}	\\
\toprule
\bottomrule
\rowcolor{gray!25}
\textbf{Qwen2.5-3B-Instruct}	&    65.24 &	52.85 &	34.76 &	27.24 	& 42.28 &	44.47 &	45.12 &	58.33 &	51.73 &	46.54 \\
\hline
Random-selection &  64.84 &	53.05 &	33.54 &	26.22 &	39.23 &	43.37 &	44.72 &	57.32 &	51.02 &	45.56  \\
MAPO  & 69.72 &	55.28 &	36.59 &	30.69 &	42.07 &	46.87 &	43.70 &	58.74 &	51.22 &	48.11  \\
LIDR  & 66.26 &	55.89 &	\textbf{49.80} &	45.53 &	51.83 &	53.86 &	53.46 &	62.20 &	57.83 &	54.99  \\
CM-Align (Ours) & \textbf{72.56} &	\textbf{64.23} &	\textbf{49.80} &	\textbf{47.36} &	\textbf{57.11} &	\textbf{58.21} &	\textbf{55.69} &	\textbf{66.06} &	\textbf{60.87} &	\textbf{58.97} 	\\

\toprule
\bottomrule
\rowcolor{gray!25}
\textbf{Llama-3-8B-Instruct}	&  55.28 & 	34.96 & 	32.11 & 	34.55 & 	38.82 & 	39.15 & 	28.86 & 	47.97 & 	38.41 & 	38.94 \\
\hline

Random-selection & 50.00 & 	27.64 & 	25.20 & 	30.69 & 	30.08 & 32.72  & 	23.98 & 	38.41 & 	31.20 & 	32.29 \\
MAPO & 60.98 & 	53.05 & 43.29 & 	41.67 & 	46.14 & 	49.02 & 	43.50 & 	55.28 & 	49.39 & 	49.13\\
LIDR & 61.38 & 	42.89 & 	35.98 & 	37.60 & 	34.55 & 	42.48 & 	38.62 & 	45.93  & 	42.28 & 	42.42\\
CM-Align (Ours)  & \textbf{64.43} & 	\textbf{58.54} & 	\textbf{48.17} & 	\textbf{44.72} & 	\textbf{50.61} & 	\textbf{53.29} & 	\textbf{48.58} & 	\textbf{58.13} & 	\textbf{53.35} & \textbf{53.31}	\\
    \toprule
    \end{tabular}
    }
    \caption{
        The pass@1 (\%) results on Multilingual HumanEval of the three models. The result in \textbf{bold} means the best result in each language.
    }
    \label{table:code-res}
\end{table*}

\subsection{Main Results}
In this section, we present the main results of our method and baselines on MATH, CODE, and General Instruction Following (GIF).

\paragraph{Results of MATH.}
The results on MGSM of Llama-3.2-3B-Instruct, Qwen2.5-3B-Instruct, and Llama-3-8B-Instruct are reported in Table \ref{table:math-res}, which demonstrate the consistent and significant superiority of our method across multiple languages and models. 
In both the ``\textit{with label}'' and ``\textit{label-free}'' settings, our CM-Align consistently achieves the highest average accuracy (\textit{All-avg}).

In the ``\textit{with label}'' setting, where \textit{En-reference} is guided by ground-truth labels, our method shows marked improvements over SFT-translation and SFT-self-rejection. 
SFT-translation only improves performance over the base models slightly, and in some cases, it even leads to a performance degradation (e.g., the \textit{All-avg} of Qwen2.5-3B-Instruct from 57.04\% to 55.72\%) due to the inevitable translationese.
Additionally, SFT-self-rejection also brings only a marginal improvement.
On the contrary, our method not only achieves the best average performance on in-domain languages but also significantly boosts performance on out-of-domain languages.

The advantages of our CM-Align are even more pronounced in the ``\textit{label-free}'' setting, which represents a more challenging and realistic application scenario. Without access to any ground-truth labels, our method consistently and substantially outperforms all \textit{label-free} baselines, including Random-selection, MAPO, and LIDR.
A key finding is that our method almost reaches the performance of the ``\textit{with label}'' setting. For example, the \textit{All-avg} 65.12\% is close to 65.84\% on Llama-3.2-3B-Instruct, and the \textit{All-avg} 61.20\% is close to 61.64\% on Llama-3-8B-Instruct.
As for other baselines, LIDR notably underperforms the original model, particularly in Chinese and French, suggesting this method heavily depends on the translation ability of the model itself, and low-quality preference data leads to performance degradation.
Random selection barely improves over the original model, confirming the need for strategic data selection methods.
While MAPO demonstrates competitive performance in the \textit{label-free} setting (60.16\% on the Llama-3-8B-Instruct model), achieving strong results for English (76.40\%) and Bengali (55.60\%), it struggles with maintaining consistent improvements across diverse languages and LLMs. 

To sum up, our method maintains strong performance for both in-domain and out-of-domain languages, which demonstrates that our method can construct high-quality preference data for better multilingual alignment. 
Furthermore, the minimal gap between ``\textit{with label}'' and ``\textit{label-free}'' settings indicates the practical utility of our approach, enabling better multilingual alignment without requiring expensive labeled data.

\begin{table*}[t]
    \centering
    \resizebox*{0.9\linewidth}{!}{
    \begin{tabular}{l|ccccc|c|cc|c|c}
    \bottomrule
  & \multicolumn{6}{c|}{\textbf{In-Domain Languages}} & \multicolumn{3}{c|}{\textbf{Out-of-Domain Languages}} & \\
\hline
\textbf{\textit{label-free}} \textbf{Methods}	&	\textbf{en}	&	\textbf{zh}	&	\textbf{es}	&	\textbf{fr}	&	\textbf{ru}	&	\quad \textbf{\textit{ID-avg}} \quad	&	\textbf{bn}	&	\textbf{de}	& 	\textbf{\textit{OOD-avg}}	& \quad	\textbf{\textit{All-avg}} \quad	\\
\hline
\rowcolor{gray!30}
\textbf{Llama-3.2-3B-Instruct}	&	64.08 &	14.65 &	27.71 &	24.48 &	14.06 &	29.00 &	47.29 &	43.22 &	45.26 &	33.64 	\\
\hline
Random-selection & 61.80 &	13.33 &	26.54 	& 25.98 &	14.14 &	28.36 &	44.58 &	42.00 &	43.29 &	32.62 	\\
MAPO	&	65.37 &	24.38 &	33.80 &	29.55 &	\textbf{17.99} &	34.22 &	52.54 &	52.73 &	52.64 &	39.48  \\
LIDR	&	\textbf{72.90} &	4.01 &	6.98 &	6.09 &	5.46 &	19.09 &	24.43 &	23.49 &	23.96 &	20.48 	\\	
CM-Align (Ours) & 63.14  &	\textbf{29.18}  &	\textbf{39.22}  &	\textbf{36.02}  &	17.71  &	\textbf{37.05}  &	\textbf{60.35}  &	\textbf{54.08} & 	\textbf{57.22}  &	\textbf{42.81} 	\\
\toprule
\bottomrule
\rowcolor{gray!25}
\textbf{Qwen2.5-3B-Instruct}	&   43.23  &		22.20  &		17.84  &		18.45  &		23.49  &		25.04  &		30.47  &		27.67  &		29.07  &		26.19 \\
\hline
Random-selection &  42.55  &	30.11  &	18.37  &	19.11  &	27.22  &	27.47  &	29.10 	 & 31.40  &	30.25  &	28.27  \\
MAPO  & 53.34 	 & 34.47  &	23.80  &	23.92 & 	30.54  &	33.21  &	30.36  &	34.09  &	32.23  &	32.93  \\
LIDR  & \textbf{62.73}  &	52.18  &	4.01  &	3.35  &	2.96  &	25.05  &	11.59  &	12.84  &	12.22  &	21.38  \\
CM-Align (Ours) & 46.37  &	\textbf{55.16}  &	\textbf{30.10}  &	\textbf{30.64}  &	\textbf{38.13}  &	\textbf{40.08}  &	\textbf{35.10}  &	\textbf{47.52}  &	\textbf{41.31}  &	\textbf{40.43} 	\\

\toprule
\bottomrule
\rowcolor{gray!25}
\textbf{Llama-3-8B-Instruct}	& 63.08  &	22.66  &	36.46  &	38.45  &	36.15  &	39.36  &	59.52  &	52.30  &	55.91  &	44.09   \\
\hline
Random-selection & 62.42  & 	26.10 	 & 38.95  & 	38.27  & 	38.42  & 	40.83  & 	65.24  & 	58.78  & 	62.01  & 	46.88  \\
MAPO  & 65.93  & 	33.48  & 	46.23  & 	40.93  & 	38.14  & 	44.94  & 	71.38  & 	61.97  & 	66.68  & 	51.15 \\
LIDR  & \textbf{70.80}  & 	28.84  & 	38.12  & 	33.97  & 	36.38  & 	41.62  & 	37.15  & 	53.29  & 	45.22  & 	42.65 \\
CM-Align (Ours) & 64.80  & 	\textbf{41.30}  & 	\textbf{48.55}  & 	\textbf{44.75}  & 	\textbf{43.51}  & 	\textbf{48.58}  & 	\textbf{81.47}  & 	\textbf{62.41}  & 	\textbf{71.94}  & 	\textbf{55.26} 	\\
    \toprule
    \end{tabular}
    }
    \caption{
        The length-controlled win rate (LCWR, \%) results on OMGEval of the three models. The baseline model for comparison is \texttt{GPT3.5-turbo}. 
    }
    \label{table:GIT-res}
\end{table*}

\paragraph{Results of CODE.}
Table \ref{table:code-res} presents the pass@1 results on Multilingual HumanEval for different alignment methods. 
The results demonstrate that our method consistently outperforms all other baselines across all languages for all three models. 
For the Llama-3.2-3B-Instruct model, our CM-Align achieves a remarkable 47.39\% average performance, representing a substantial improvement of nearly 18 percentage points over the original model (29.44\%). 
The improvements are particularly pronounced in English (53.66\% vs. 35.37\%) and Chinese (52.03\% vs. 30.89\%). 
Similarly, for Qwen2.5-3B-Instruct and Llama-3-8B-Instruct models, our CM-Align reaches 58.97\% and 53.31\%, significantly outperforming the original models and other baselines. 
LIDR exhibits inconsistent performance, struggling significantly with the Llama-3.2-3B-Instruct model (16.11\% average) but performing more competitively with Qwen2.5-3B-Instruct (54.99\%), which further indicates that the performance of LIDR depends on the translation ability of the original model.
Notably, our CM-Align demonstrates strong generalization capabilities, with substantial improvements in out-of-domain languages. 
These results highlight the quality of our constructed preference data, which can significantly improve code generation capabilities across diverse languages.

\paragraph{Results of GIF.}
Table \ref{table:GIT-res} presents the evaluation results on the OMGEval benchmark, comparing the performance against \texttt{GPT-3.5-turbo}. 
We observe several important trends across both model sizes. 
First, our method CM-Align consistently achieves the best overall performance compared to other baselines. 
While LIDR excels specifically in English (achieving 72.90\%, 62.73\%, and 70.80\% win rates), it struggles considerably with non-English languages. 
We guess that the preference data constructed by LIDR is more suitable for improving the English general instruction following capacity, in which scenario, the rewards are more accurate.
Additionally, MAPO shows balanced improvements across languages, but doesn't match our method. 
Our method also demonstrates exceptional generalization to out-of-domain languages, with substantial improvements in Bengali and German.
All these results prove the effectiveness of our method again.

\begin{table}[t]
    \centering
    \resizebox*{\linewidth}{!}{
    \begin{tabular}{l|ccc}
    \bottomrule
    \textbf{MATH} & \textbf{Llama-3.2-3B} & \textbf{Llama-3-8B} & \textbf{Qwen2.5-3B} \\
    \hline
    Ours (\textit{with label}) & 100\%  & 100\% & 100\% \\
    \hline
    Random-selection & 13.94\%  & 12.78\% & 9.08\% \\
    MAPO & 6.69\%  & 17.19\% & 12.90\% \\
    LIDR & 19.64\%  & 17.47\% & 15.31\% \\
    Ours (\textit{label-free}) & \textbf{91.67\%}  & \textbf{78.85\%} & \textbf{89.16\%} \\
    \toprule
    \end{tabular}
    }
    \caption{
        The average reward accuracy of the 5 in-domain languages.
    }
    \label{table:reward-acc}
\end{table}

\section{Analysis}

\subsection{Accuracy of Rewards}
For the MATH task, we define reward accuracy based on the alignment between selected examples and ground truth. 
Specifically, an accurate reward occurs when the chosen example's final answer matches the ground truth, while the rejected example's answer differs from it. 
We show the average reward accuracy of the 5 in-domain languages across different methods and models in Table \ref{table:reward-acc}, which demonstrates that our method successfully constructs high-quality preference data with precise reward signals for different models.

\begin{table}[t]
    \centering
    \resizebox*{\linewidth}{!}{
    \begin{tabular}{l|c|ccc}
    \bottomrule
    \textbf{MATH} & \textbf{Accuracy} & \textbf{\textit{ID-avg} } & \textbf{\textit{OOD-avg}} & \textbf{\textit{All-avg} } \\
    \hline
    Ground-truth & 100\% & 72.32\% & 59.36\% & 65.84\% \\
    \hline
    Ours  & \textbf{90.84\%}  & \textbf{70.24\%} & \textbf{60.00\%} & \textbf{65.12\%} \\
    Random-En  & 80.96\% & 67.28\% & 56.80\% & 62.04\%\\
    \toprule
    \bottomrule
    \textbf{CODE} & \textbf{/} & \textbf{\textit{ID-avg} } & \textbf{\textit{OOD-avg}} & \textbf{\textit{All-avg} } \\
    \hline
    Ours  & / & \textbf{45.89\%} & \textbf{51.12\%}  & \textbf{47.39\%} \\
    Random-En  & / & 43.50\% & 43.79\% & 43.47\%\\
    \toprule
    \bottomrule
    \textbf{GIF} & \textbf{/} & \textbf{\textit{ID-avg} } & \textbf{\textit{OOD-avg}} & \textbf{\textit{All-avg} } \\
    \hline
    Ours  & / & \textbf{37.05\%}  & \textbf{57.22\%} & \textbf{42.81\%} \\
    Random-En  & / & 32.06\% & 56.55\% & 39.05\%\\
    \toprule
    \end{tabular}
    }
    \caption{
        The performance degradation of utilizing randomly selected \textit{En-reference} to construct multilingual preference data.
        ``Accuracy'' means the accuracy of \textit{En-reference} compared to the ground-truth label (only for MATH), and ``\textit{*-avg}'' represents the average results of Llama-3.2-3B-Instruct.
    }
    \label{table:dif-en-references}
\end{table}

\subsection{Ablation}
In this section, we investigate the effectiveness of the consistency-guided selection strategy for \textit{En-reference} and the necessity of designing task-specific consistency metrics for constructing the preference data.

\paragraph{Random \textit{En-reference}.}
We list the performance of using different \textit{En-reference} in Table \ref{table:dif-en-references}.
For MATH, randomly selecting one response as \textit{En-reference} only has 80.96\% accuracy, 
while our consistency-guided selection (voting for Math) strategy can improve the accuracy to 90.84\%.
The higher accuracy of \textit{En-reference}, the higher accuracy of multilingual preference data.
As a result, our method exhibits better performance on the MGSM benchmark in both in-domain and out-of-domain languages compared to the ``Random-En'' setting and achieves approximate performance with the setting of ``Ground-truth'' as \textit{En-reference}.
For CODE and GIF, our method also outperforms ``Random-En''.
In conclusion, these results demonstrate the effectiveness of our consistency-guided English reference selection strategy for selecting a reliable English anchor to improve multilingual alignment.

\paragraph{Unify consistency metric for different tasks.}
We list the results of MATH/CODE tasks with the embedding-based metric (\textit{i.e.}, $\mathrm{Cons}_{\text{GIF}}$) in Table \ref{table:embedding-metric}.
The results show that even if the \textit{En-reference} is reliable, utilizing an unsuitable metric $\mathrm{Cons}_{\text{GIF}}$ can not select accurate chosen/rejected samples for MATH/CODE to construct high-quality multilingual preference data, resulting in poor alignment performance.
Overall, these results prove the necessity and effectiveness of our designed task-specific consistency metrics.

\begin{table}[t]
    \centering
    \resizebox*{\linewidth}{!}{
    \begin{tabular}{l|ccc}
    \bottomrule
    \textbf{Criteria} & \textbf{\textit{ID-avg} } & \textbf{\textit{OOD-avg}} & \textbf{\textit{All-avg} } \\
    \hline
    MATH + $\mathrm{Cons}_{\text{math}}$ (Ours) & \textbf{70.24\%} & \textbf{60.00\%} & \textbf{65.12\%} \\
    MATH + $\mathrm{Cons}_{\text{GIF}}$ & 62.88\% & 52.64\% & 57.76\% \\
    \hline
    CODE + $\mathrm{Cons}_{\text{code}}$ (Ours) & \textbf{45.89\%} & \textbf{51.12\%} & \textbf{47.39\%} \\
    CODE + $\mathrm{Cons}_{\text{GIF}}$ & 35.16\% & 34.96\% & 35.10\% \\
    
    \toprule
    \end{tabular}
    }
    \caption{
        The results of MATH/CODE with the embedding-based metric ($\mathrm{Cons}_{\text{GIF}}$) for Llama-3.2-3B-Instruct.
    }
    \label{table:embedding-metric}
\end{table}

\subsection{Performance Improvement of Scaling Models}
\begin{figure}[t]
    \centering
    \includegraphics[width=\linewidth]{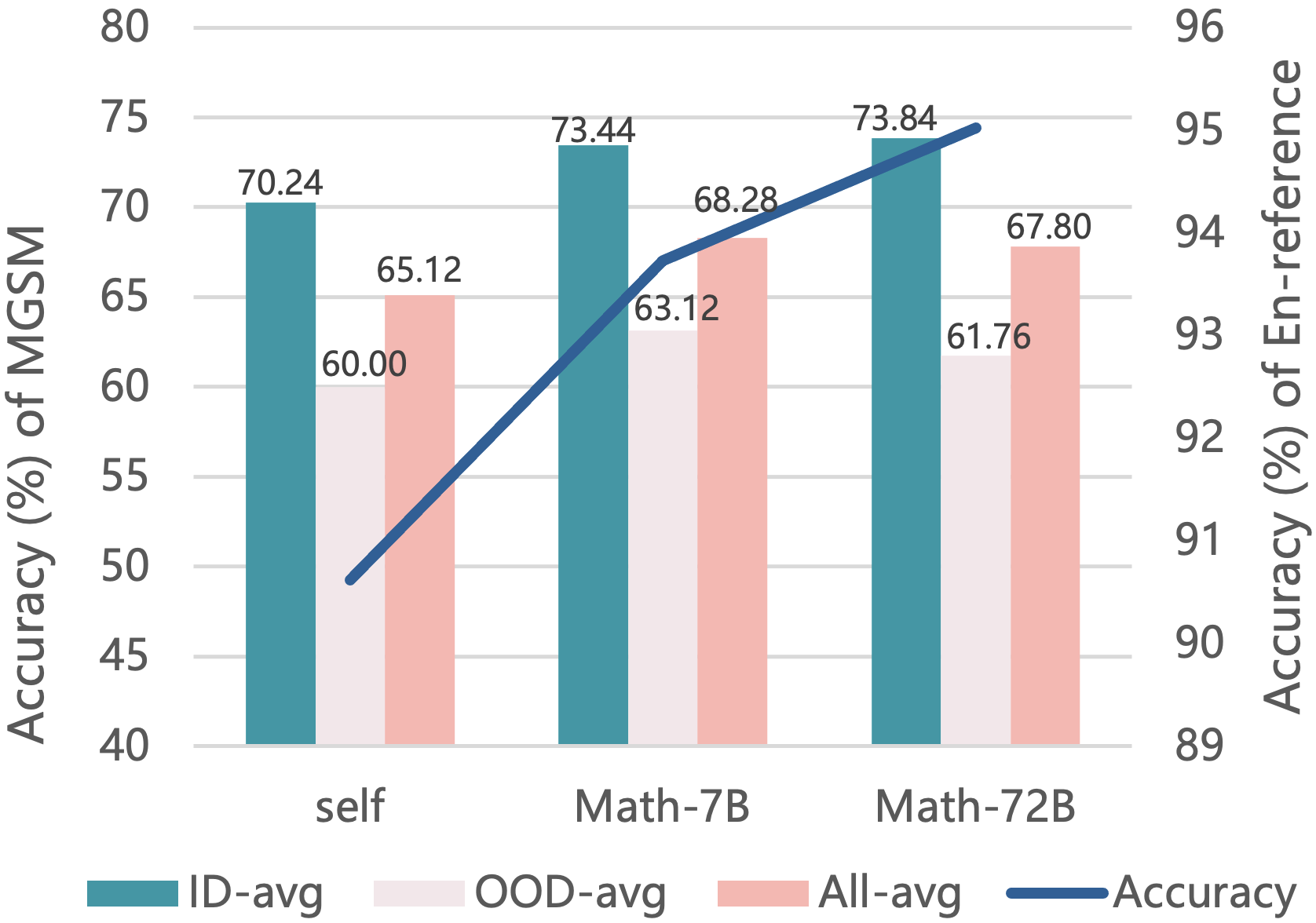}
    \caption{The performance improvement with model size scaling. ``Accuracy'' (right axis) denotes the accuracy of \textit{En-reference} and ``\textit{*-avg}'' (left axis) represents the average results on MGSM of Llama-3.2-3B-Instruct. ``self/Math-7B/72B'' means utilizing the Llama-3.2-3B-Instruct/Qwen2.5-Math-7B/72B-Instruct model for generating responses.}
    \label{fig:scaling_models}
\end{figure}

In this section, we explore whether the multilingual alignment performance improves when scaling the model sizes for generating responses.
Specifically, we utilize the Qwen2.5-Math-7B/72B-Instruct \cite{yang2024qwen25mathtechnicalreportmathematical} model to generate responses to each question and conduct our method.
The results in Figure \ref{fig:scaling_models} show that the accuracy of \textit{En-reference} is higher as the model size larger (please refer to the blue line), and the results on MGSM also improve, \textit{e.g.}, ``ID-avg'' from 70.24 to 73.44/73.84, and ``All-avg'' from  65.12 to 68.28/67.80.
However, the average performance does not improve consistently when the model size scales from 7B to 72B.
We suspect that with constrained data volumes (\textit{i.e.}, up to 4.5K samples per language), there exists a performance ceiling that cannot be overcome merely by scaling model sizes. 
To achieve better multilingual alignment results, the primary focus may be shifted towards expanding the size of the training dataset.




\section{Conclusion}
In this paper, we design a consistency-based data selection method to construct high-quality multilingual preference data for improving multilingual alignment.
Specifically, our method includes two procedures: consistency-guided English reference selection and cross-lingual consistency-based preference data construction.
We conduct extensive experiments on three LLMs with different model sizes and three common tasks (Math, Code, and General Instruction Following).
The experimental results demonstrate the effectiveness and superiority of our CM-Align.

\section*{Limitations}
To conduct a fast experiment, we do not adopt Iterative DPO, although this training strategy has been proven effective in MAPO \cite{she-etal-2024-mapo} and LIDR \cite{yang2025languageimbalancedrivenrewarding}.
In the future, we will explore the performance improvement when integrating the Iterative DPO training.
Additionally, further experimental investigation is needed to determine whether expanding the per-task training data volume would lead to enhanced multilingual alignment performance.

\section*{Acknowledgments}
The research work described in this paper has been supported by the National Natural Science Foundation of China (No. 62476023, 61976016, 62376019, 61976015), and the authors would like to thank the anonymous reviewers for their valuable comments and suggestions to improve this paper.

\bibliography{custom}

\appendix
\onecolumn
\newpage

\section{Extraction Regular Expressions for Math responses}
\label{sec:appendix-regx_math}
For the responses to Math prompts, we first utilize the following code to extract the final numerical value as the answer for judging the consistency of any two responses.
\lstset{
    numbers=none
}
\begin{lstlisting}
def extract_last_num(text: str) -> float:
    text = re.sub(r"(\d),(\d)", "\g<1>\g<2>", text)  # processing for 123,456
    res = re.findall(r"(\d+(\.\d+)?)", text)  # matching for 123456.789
    if len(res) > 0:
        num_str = res[-1][0]
        return float(num_str)
    else:
        return 0.0
\end{lstlisting}

\section{Codes for Normalizing Code Snippets}
\label{sec:appendix-codes_for_code}
For the responses to Code prompts, we first extract the code snippet and utilize the following code to normalize the code snippet.
\lstset{
    numbers=none
}
\begin{lstlisting}
class CodeNormalizer:
    def __init__(self, 
                 remove_comments=True,
                 anonymize_variables=True,
                 standardize_format=True):
        self.remove_comments = remove_comments
        self.anonymize_variables = anonymize_variables
        self.standardize_format = standardize_format

    def process(self, code):
        """Carry out the complete code normalization process."""
        if self.remove_comments:
            code = self._remove_comments(code)
        if self.anonymize_variables:
            code = self._anonymize_variables(code)
        if self.standardize_format:
            code = self._standardize_format(code)
        return code

    def _remove_comments(self, code):
        """Remove all comments (including inline comments) and retain the code structure."""
        try:
            io_obj = StringIO(code)
            tokens = list(tokenize.generate_tokens(io_obj.readline))
            filtered_tokens = [t for t in tokens if t.type != tokenize.COMMENT]
            new_code = tokenize.untokenize(filtered_tokens)
            return new_code.decode('utf-8').replace('\r\n', '\n').replace('\r', '\n')
        except Exception as e:
            print(f"_remove_comments error: {e}")
            return code

    def _anonymize_variables(self, code):
        """Variable anonymization for maintaining scope consistency."""
        class Renamer(ast.NodeTransformer):
            def __init__(self):
                self.var_map = {}
                self.counter = 0

            def visit_Name(self, node):
                if isinstance(node.ctx, ast.Store):
                    if node.id not in self.var_map:
                        self.var_map[node.id] = f"var{self.counter}"
                        self.counter += 1
                if node.id in self.var_map:
                    return ast.Name(id=self.var_map[node.id], ctx=node.ctx)
                return node
        try:
            tree = ast.parse(code)
            tree = Renamer().visit(tree)
            return astor.to_source(tree)
        except Exception as e:
            print(f"_anonymize_variables error: {e}")
            return code

    def _standardize_format(self, code):
        """Code format standardization."""
        try:
            import black
            return black.format_str(code, mode=black.FileMode())
        except ImportError:
            try:
                tree = ast.parse(code)
                return astor.to_source(tree)
            except:
                return code
        except Exception as e:
            print(f"_standardize_format error: {e}")
            return code
\end{lstlisting}

\section{Training Details}\label{appendix:training_details}
The other training details are listed in Table \ref{tab:train_config}, which are determined according to the best performance. Some of the settings follow previous work \cite{yang2025languageimbalancedrivenrewarding, li2025multilingualcollaborativedefenselarge, zhang2024dualspaceknowledgedistillationlarge, zhang-etal-2025-multilingual, zhang-etal-2023-quality}.

\begin{table}[h]
    \centering
    \resizebox{0.5\linewidth}{!}{
        \begin{tabular}{c|cc|cc|cc}
            \bottomrule
            \multirow{2}{*}{\textbf{Settings}} & \multicolumn{2}{c|}{\textbf{Math}} & \multicolumn{2}{c|}{\textbf{Code}} &\multicolumn{2}{c}{\textbf{GIF}} \\
            \cline{2-7}
            & 3B & 8B & 3B & 8B & 3B & 8B \\ 
            \hline
            $\beta$ & 0.1 & 0.1 & 0.1 & 0.1 & 0.1 & 0.1 \\
            $\gamma$ & 1.0 & 1.0 & 0.1 & 0.1 & 0.0 & 0.0 \\
            Batch Size & 256 & 256 & 64 & 64 & 16 & 16 \\
            Learning Rate & 1e-5 & 1e-6 & 1e-6 & 1e-6 & 1e-6 & 5e-7\\
            Epoch & 3 & 3 & 1 & 1 & 1 & 1\\
            
            \toprule
        \end{tabular}
    }
    \caption{Detailed training configurations for the three models and three tasks. ``3B'' denotes Llama-3.2-3B-Instruct and Qwen2.5-3B-Instruct, and ``8B'' denotes Llama-3-8B-Instruct.}\label{tab:train_config}
\end{table}

\end{document}